\documentclass[10pt,twocolumn,letterpaper]{article}

\usepackage{iccv}

\usepackage{graphicx}
\usepackage{amsmath}
\usepackage{amssymb}
\usepackage{booktabs}

\usepackage{multirow}
\usepackage{booktabs}
\usepackage[table,xcdraw]{xcolor}
\usepackage{colortbl}
\usepackage{url}
\usepackage{rotating}
\usepackage{booktabs}
\usepackage{multirow}
\usepackage{soul}
\makeatletter
\@namedef{ver@everyshi.sty}{}
\makeatother
\usepackage{tikz}
\usepackage{comment}
\usepackage{amsmath,amssymb} 
\usepackage{color}
\usepackage{pifont}
\usepackage{bm}  

\usepackage{wrapfig,lipsum,booktabs}

\usepackage[accsupp]{axessibility}  
\newcommand{\cmark}{\ding{51}}%
\newcommand{\xmark}{\ding{55}}%

\usepackage{times}
\usepackage{epsfig}
\usepackage{graphicx}
\usepackage{amsmath}
\usepackage{amssymb}

\usepackage{algorithm}
\usepackage{bm}
\usepackage{algorithmic}
\usepackage{listings}
\usepackage{makecell}  
\usepackage{diagbox} 
\usepackage{mathtools} 
\usepackage{multirow}

\usepackage[table]{xcolor}
\usepackage[pagebackref, breaklinks=true, colorlinks, citecolor=citecolor, linkcolor=linkcolor, bookmarks=false]{hyperref}
\definecolor{citecolor}{HTML}{0071BC}
\definecolor{linkcolor}{HTML}{ED1C24}

\newcommand{\FX}[1]{{{\color{black}#1}}}

\definecolor{bestcolor}{gray}{.9}

\newcommand{\modelname}{Multi-Modality Prompt Meta-Learning}
\newcommand{\shortmodelname}{MUPPET}

\iccvfinalcopy 

\def\iccvPaperID{1246} 

\ificcvfinal\fi

\begin{document}

\title{Multi-Modal Few-Shot Temporal Action Detection}

\author{Sauradip Nag$^{1,5}$
\and
Mengmeng Xu$^{2,4}$\thanks{This work was done while internship in Meta,UK}
\and
Xiatian Zhu$^{1,3}$
\and
Juan-Manuel P\'erez-R\'ua$^{2}$
\and \newline
Bernard Ghanem$^{4}$
\and 
Yi-Zhe Song$^{1,5}$
\and 
Tao Xiang$^{1,5}$ 
\and \newline
{\small $^1$ CVSSP, University of Surrey, UK} ~ 
{\small $^2$ Meta, UK} ~
{\small $^3$ Surrey Institute for People-Centred Artificial Intelligence, UK} \\
{\small $^4$ KAUST, Saudi Arabia} ~
{\small $^5$ iFlyTek-Surrey Joint Research Center on Artificial Intelligence, UK}
}

\maketitle
\ificcvfinal\thispagestyle{empty}\fi

\begin{abstract}
Few-shot (FS) and zero-shot (ZS) learning are two approaches for scaling temporal action detection (TAD) to new classes.  
The former adapts a pretrained vision model to a new task represented by as few as a single video per class, whilst the latter requires no training examples by exploiting a semantic description of the new class.
In this work, we introduce a new {\em multi-modality few-shot} (MMFS) TAD problem, as a marriage of FS-TAD and ZS-TAD by leveraging few-shot support videos and a new class name jointly.
To tackle this problem,
we further introduce a novel {\bf\em MUlti-modality PromPt mETa-learning} (\shortmodelname) method.
{This is enabled by efficiently bridging pretrained vision and language models whilst maximally reusing already learned capacity.
Concretely, we construct multi-modal prompts by mapping support videos into the textual token space of a vision-language model using a meta-learned adapter-equipped visual semantics tokenizer.}
To tackle large intra-class variation, we further design a query feature regulation scheme. 
Extensive experiments on ActivityNetv1.3 and THUMOS14
demonstrate that our \shortmodelname{} outperforms state-of-the-art alternative methods, often by a large margin.
{\shortmodelname{} can be easily extended to few-shot object detection, 
achieving new state-of-the-art on MS-COCO. The code will be made available in \href{https://github.com/sauradip/MUPPET}{https://github.com/sauradip/MUPPET}}

\end{abstract}

\begin{figure}[t]
    \centering
    \includegraphics[scale=0.4]{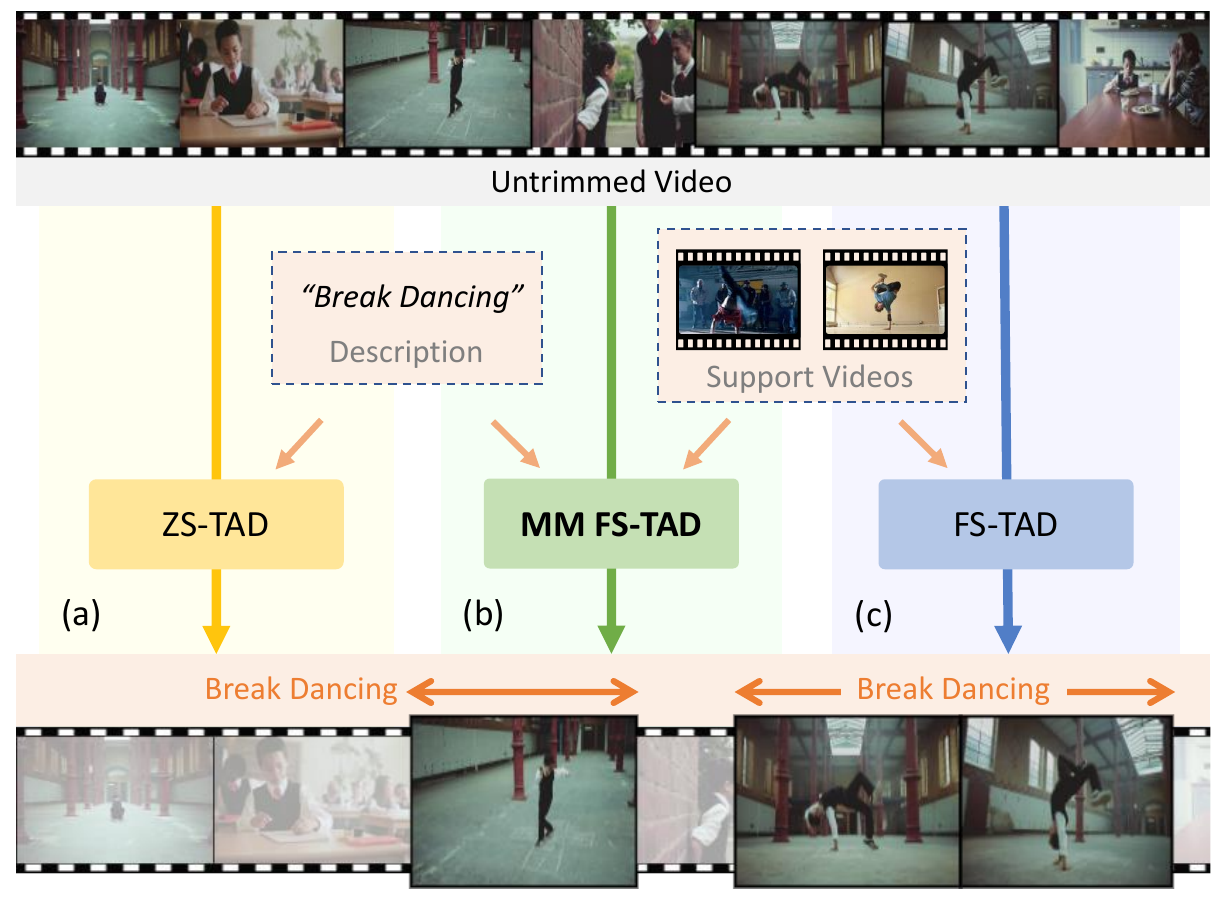}
     \vspace{-0.15in}
    \caption{\textbf{Illustration of different problem settings}. (a) \textit{Zero-shot temporal action detection} (ZS-TAD) 
    represents a new class using a semantic description of its name (\ie, textual input).
    (c) \textit{Few-shot temporal action detection} (FS-TAD)
    can learn a new class from a few support (training) videos (\ie, visual input).
    (b) Our new \textit{multimodal few-shot temporal action detection} (MMFS-TAD) can leverage both textual and visual inputs.
    }
    \label{fig:problem}
    \vspace{-0.2in}
\end{figure}
\vspace{-0.2in}
\section{Introduction}
The objective of temporal action detection (TAD) is to predict the temporal duration (\ie, start and end time) and the class label of each action instance in an untrimmed video \cite{idrees2017thumos,caba2015activitynet}.
Conventional TAD methods \cite{xu2021low,xu2020g,buch2017sst,wang2017untrimmednets,zhao2017temporal,nag2022gsm,nag2021temporal} are based on supervised learning, requiring  many (\eg, hundreds) videos per class with costly segment-level annotations for training.  This thus severely limits their ability to scale to many classes.  
To alleviate this problem, \texttt{few-shot} (FS) \cite{yang2018one,yang2020localizing,zhang2020metal,nag2021few} and \texttt{zero-shot} (ZS) \cite{zhang2020metal,ju2021prompting,nag2022pclfm} learning based TAD methods have been recently introduced.

Specifically, FS-TAD aims to learn a model that can adapt to new action classes
with as few as a single training video per class (Fig.~\ref{fig:problem}(c)).
This is achieved often by meta-learning a TAD model over a distribution of simulated tasks on seen classes.
ZS-TAD further removes the need for any training samples from new classes. Instead, new classes are represented by projecting their class names into some semantic space (\eg, attributes, word embeddings), (Fig.~\ref{fig:problem}(a)). Once the semantic space is aligned with a visual feature space, a model trained on seen classes can be applied/transferred to new ones. 
The recent emergence of large-scale Visual-Language (ViL) models (\eg, CLIP \cite{radford2021learning} and ALIGN \cite{jia2021scaling}) have clearly advanced  the research of {\em zero-shot learning} in general, and ZS-TAD in particular \cite{ju2021prompting,nag2022pclfm}. This is because these ViL models offer a strong alignment between the text (\eg, action class name or description) and visual  (\eg, video content features) modalities, which is a key requirement for ZS-TAD.

FS-TAD and ZS-TAD have thus far been studied {\em independently}. This is in contrast to the object classification/detection domain where attempts have been made to unify the two problems in a single framework \cite{han2022multimodal,fu2015transductive}. Tackling them jointly in TAD makes sense for a number of reasons. (1) These two problems have a  shared goal of scaling up TAD. (2) In practice, one often has a few examples of a new class, together with the class name. (2) The two tasks are complementary to one another, \eg, the semantic description extracted from an action name can compensate for the limitation of few-shot examples in representing the large intra-class variations in action. One of the potential obstacles with this unification is that the semantic description from an action name is often too weak to describe the rich visual context of that action, and therefore cannot match with the representation power of even a single video example \cite{awad2016trecvid}. However, this has been changed with the ever-stronger ViL models available, as mentioned above.

In this work, for the first time, TAD is studied under a new setting, namely
\texttt{multimodal few-shot temporal action detection} (MMFS-TAD),
characterized by learning from both support videos (\ie, visual modality) and class names (\ie, textual modality).
More specifically, we introduce a novel
{\em \modelname} ({\bf \shortmodelname}) method  to efficiently fuse few-shot visual examples and high-level class semantic text information.
Grounded on a pre-trained vision-language (ViL) model (\eg, CLIP), {\shortmodelname}  integrates meta-learning with learning-to-prompt
in a unified TAD framework.
This is made possible by introducing a {\em multimodal prompt learning module} that maps 
the support videos of a novel task
into the textual token space of the ViL model using a meta-learned adapter-equipped visual semantics tokenizer.
\textcolor{black}{During meta-training, 
this tokenizer is jointly learned with other components in order to 
map 
visual representations into a designated $<context>$ token compatible with the language model. 
During inference (\ie, meta-test), 
given a new task, our model can induce 
by digesting few training examples and class names in a data-driven manner.}
With the ViL's text encoder, our multimodal prompt can be then transformed 
to multimodal class prototypes for action detection.
To tackle the large intra-class challenge due to limited support samples,
we design a query feature regulation strategy
by meta-learning a masking representation from the support sets and attentive conditioning. 

We summarize our {\bf contributions} as follows.
{\bf (1)} We propose the {multimodal few-shot temporal action detection} (MMFS-TAD) problem.
{\bf (2)} To solve this new problem, we introduce a novel {\em \modelname} (\shortmodelname) method that integrates meta-learning and learning-to-prompt
in a single formulation. It can be easily plugged into existing TAD architectures and is flexible in tackling FS-TAD and ZS-TAD either independently or jointly.
{\bf (3)} To better relate query videos with limited support samples, we design a query feature regulation scheme based on meta-learning a masking representation from the support sets, and attentive conditioning.
%
{\bf (4)} 
Extensive experiments on ActivityNet-v1.3 and THUMOS14
validate the superiority of our \shortmodelname{} over state-of-the-art alternative methods in 
the MMFS-TAD, ZS-TAD, and FS-TAD settings.
Under minimal adaptation, \shortmodelname{} can also achieve superior few-shot object detection performance on COCO.

\section{Related Works}
\noindent \textbf{Temporal action detection}
Substantial progress has been made in TAD. Inspired by object detection \cite{ren2016faster},
R-C3D \cite{xu2017r} uses anchor boxes in the pipeline of {proposal generation and classification}.
Similarly, TURN \cite{gao2017turn} aggregates local features to represent snippet features for temporal boundary regression and classification. SSN \cite{zhao2017temporal} decomposes an action instance into start:course:end
and employs structured temporal pyramid pooling
for proposal generation.
BSN \cite{lin2018bsn} generates proposals with high start and end probabilities by modeling the start, end, and actionness at each time. Later, BMN \cite{lin2019bmn} improves the actionness by generating a boundary-matching confidence map. 
For better proposal generation, G-TAD \cite{xu2020g}
learns semantic and temporal context via graph convolutional networks. 
CSA \cite{sridhar2021class} enriches the proposal temporal context via attention transfer. 
%
Unlike most previous models 
adopting a {sequential}
localization and classification pipeline,
TAGS \cite{nag2022gsm} introduces a different design
with parallel localization and classification
based on a notion of global segmentation masking.
All the above methods are supervised
with reliance on large training data, and thus less scalable.

\noindent \textbf{Few-shot temporal action detection }
By fast adaptation of a model to any given new class with few training samples,
few-shot learning (FSL) provides a solution for scalability \cite{vinyals2016matching, sung2018learning, snell2017prototypical}.
FSL is often realized by meta-learning which simulates new tasks with novel classes represented by only a handful of labeled samples.
FSL has been introduced to TAD in 
\cite{yang2018one}, by
incorporating sliding window in a matching network \cite{vinyals2016matching} strategy.
Later on, \cite{zhang2020metal} consider weak video-level annotation of untrimmed training videos.
\cite{yang2021few} performed few-shot spatio-temporal action detection with a focus on a single new class at a time. Recently, \cite{nag2021few} used the Transformer for adapting the support learned features to the query features in untrimmed videos.
\begin{figure*}[h]
    \centering
    \includegraphics[scale=0.24]
    {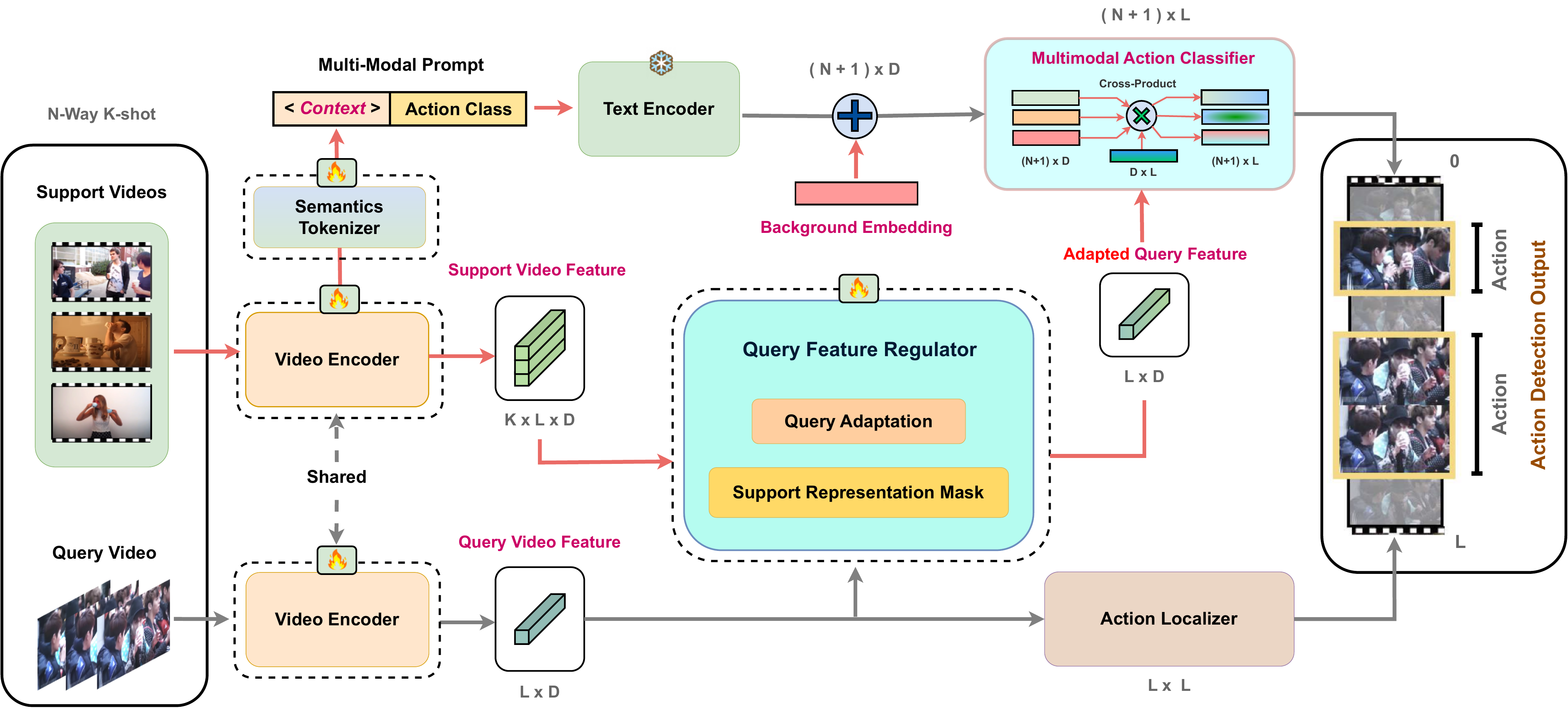}
    \caption{\textbf{Overview of our {\em \modelname} (\shortmodelname) method}. We adopt the global mask-based TAD architecture \cite{nag2022gsm,nag2022pclfm}. 
    The key components of \shortmodelname{}
    include (1) multimodal prompt meta-learning
    (Sec.~\ref{sec:mm_prompt}) and
    (2) query feature regulation (Sec.~\ref{sec:query_reg}). Dotted-line boxes represent the meta-learned modules.
    }
    \label{fig:overview}
      \vspace{-0.25in}
\end{figure*}

\noindent \textbf{Zero-shot temporal action detection }
Alternatively, zero-shot learning allows for recognizing new classes with no labeled training data.
This line of research has advanced significantly due to the 
promising power of large vision-language (ViL) models,
\eg, CLIP trained by 400 million image-text pairs \cite{radford2021learning}.
Follow-ups further boost the zero-shot transferable ability, \eg, CoOp \cite{zhou2021learning}, 
CLIP-Adapter \cite{gao2021clip}, and
Tip-adapter \cite{zhang2021tip}.
In video domains, a similar idea has also been explored for transferable representation learning \cite{miech2020end}, and text-based action localization \cite{paul2021text}. 
CLIP is used recently in action recognition \cite{wang2021actionclip} and TAD \cite{ju2021prompting,nag2022pclfm}. 

In this work, FS-TAD and ZS-TAD are unified 
in a new MMFS-TAD setting. 
Our \shortmodelname{} can tackle either FS-TAD or ZS-TAD, and crucially both simultaneously when both visual examples and class descriptions are available. 



\vspace{-0.10in}
\section{Methodology}


\subsection{Preliminaries: Multi-Modal Few-Shot }

\textcolor{black}{We establish the proposed multimodal few-shot learning by integrating class semantic information (\eg, text such as action class names) to few-shot learning \cite{dong2018few}.
To facilitate understanding, we follow the standard episode-based meta-learning convention.
Given a new task in each episode}
%
%
with a few labeled support videos per unseen class (\ie, visual modality) and class names (\ie, textual modality),
we aim to learn a TAD model for that task.  \textcolor{black}{For a $N$-way $K$-shot setting, the support set $\textit{S}$ 
consists of
$K$ labeled samples for each of the $N$
action classes. 
The query set $Q$ has a single sample per class. Key to MMFS-TAD is to leverage limited video examples
and action class names jointly.}
We have a base class set $C_{base}$ for
training, and a novel class set $C_{novel}$ for test.
For testing cross-class generalization,
we ensure they are disjoint: $C_{base} \bigcap C_{novel} = \phi$.
The base and novel sets are denoted as 
$D_{base} = \left \{ \left ( V_{i}, Y_{i} \right ), Y_{i} \in C_{base}\right \}$ and $D_{novel} = \left \{ \left ( V_{i}, Y_{i} \right ) , Y_{i} \in C_{novel}\right \}$ respectively.
Under the proposed setting, each training video $V_{i}$ is associated with
segment-level annotation 
$Y_{i} = \left \{ (s_{t},e_{t},c), t \in \{1,..,M\}, c \in \textit{C}\right \}$ including $M$ segment labels each with the start and end time and action class $c$. 
%
 In evaluation, 
for each task, we randomly sample a set of classes $\textit{L} \sim C_{novel} $ each with 
the support set \textit{S} ($K$ videos) and the query set \textit{Q} (one video) respectively. 
The labels of $S$ are accessible for few-shot learning
whilst that of $Q$ is only used for performance evaluation.

\begin{figure*}[h]
    \centering
    \includegraphics[scale=0.43]{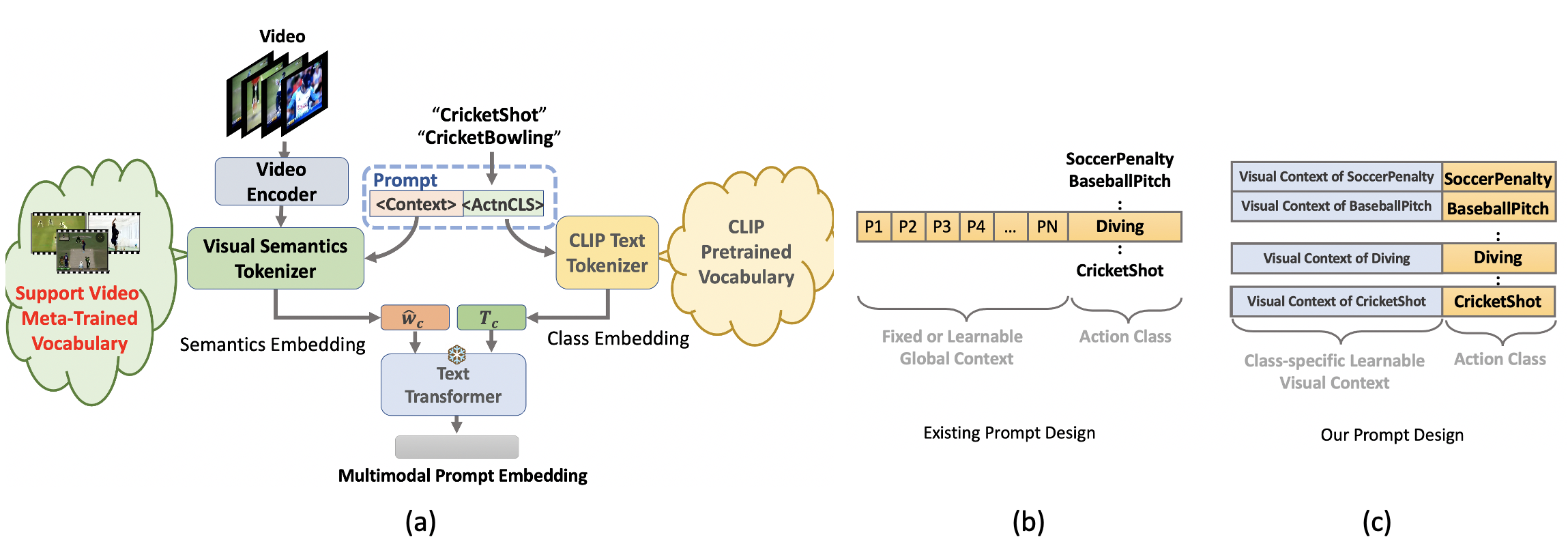}
    \vspace{-0.10in}
    \caption{\textbf{Multimodal prompt meta-learning}. (a) We meta-learn a visual semantics tokenizer for translating the support videos (\ie, visual modality) to the textual token space of a pretrained ViL model.
    Together with the tokens of class names, this mapping facilitates the creation of multimodal prompts using the pretrained text encoder. 
    (b) Unlike previous class-generic visual prompts, we consider more discriminative class-specific counterparts.}
    \label{fig:mmprompt}
    \vspace{-0.20in}
\end{figure*}

\textcolor{black}{As illustrated in Fig.~\ref{fig:overview}, our {\em \modelname} (\shortmodelname) model consists of a  video encoder, a meta-mapper, a text encoder, a query regulator and decoder heads. 
}

\subsection{Multi-Modal Prompt Learning}
\label{sec:mm_prompt}
\FX{Given a few training videos with a class name, we design a meta-learning process
for multi-modal fusion in prompts.
As shown in Fig.~\ref{fig:mmprompt}, we extract each video's feature with a video encoder and project it into the textual token space with a visual semantics tokenizer. 
The class name is then encoded by the text tokenizer~\cite{gao2021clip} and further concatenated to the visual tokens. We apply a text encoder to create the multi-modal prompt embedding. 
Let us describe the video encoder, visual semantics tokenizer, and text encoder next.} 

\paragraph{Video encoder}
\textcolor{black}{
\FX{We employ a residual adapter $\theta$ on a pretrained Vision-Transformer (ViT)~\cite{chen2022adaptformer} to learn the video priors. Since the training data in TAD is relatively small due to costly labeling, using adapter facilitates the reuse of pretrained knowledge during fine-tuning.}
%
%
Formally, an untrimmed video is encoded into a feature tensor $V \in \mathbb{R}^{ 3 \times T \times H \times W}$ with $T$
frames and a spatial resolution of $H \times W$. 
To capture global context information useful for modeling long-term dependency, another light transformer $\phi$ is applied to the time dimension:
\begin{align}
    F = \mathbb{V}(V;[\theta,\phi]) \in \mathbb{R}^{t \times D},
\end{align}
where $D$ is the snippet feature dimension and $t$ is the number of temporal snippets. \FX{Then,} following \cite{lin2019bmn,xu2020gtad},
we uniformly sample $L$ equidistant \FX{features} over all the $t$ snippets to obtain the video feature $E \in \mathbb{R}^{D \times L}$. \FX{Finally},
given a $K$-shot task $T_{i} = \{S_{i}, Q_{i}\}$,
we extract the support features $E_{s} \in \mathbb{R}^{K \times D \times L}$ for $S_i$ and query features $E_{q} \in \mathbb{R}^{D \times L}$ for $Q_i$. \FX{More details are given in Supplementary.}
}




\noindent\textbf{Visual semantics tokenizer} \textcolor{black}{
%
\FX{Multi-modal learning requires the interaction across modalities in a common space. To that end,}
we \FX{propose} a visual semantics tokenizer $f_{\theta_{s}}$ based on set Transformer \cite{lee2019set} as:
\begin{align}
\label{eq:3}
    \hat{w}_{c} = f_{\theta_{s}}(\hat{E}_{s}^k | k=1,2,...K)  
    \in W,
\end{align}
where $\theta_{s}$ is the parameters, and $\hat{E}_{s}$ is the action/foreground feature of support videos\FX{, which are} obtained by trimming off the background frame features of ${E}_{s}$.
$W$ is the textual token space. 
\FX{Concretely,} we set the query/key/value of $f_{\theta_{s}}$ all to $\hat{E}_{s}^k$ 
\FX{so that} the token $\hat{w}_{c}$ predicted by  the visual feature 
\wrt{} the designated learnable $<context>$ token (see Fig.~\ref{fig:mmprompt}(a)) is learned to be compatible with the textual embeddings in $W$. 
}

\noindent \textbf{Text encoder} \textcolor{black}{
The text encoder processes a text sequence
and outputs a feature distribution.
It has a text tokenizer $\psi \in \mathbb{R}^D$ 
that maps each token of the input into a representation space. Instead of learning a common prompt embedding for all the target classes as \cite{zhou2021learning} (Fig.~\ref{fig:mmprompt}(b)), we propose to learn a {\em class-specific context embedding} (Fig.~\ref{fig:mmprompt}(c)) for generating more discriminative 
new class representation (Table \ref{tab:prompt}). We obtain the class token embeddings $T_{c} = \psi(c)$ using the pretrained language model tokenizer,
then concatenate it with the $<context>$ token embedding $\hat{w}_{c}$ from the visual semantics tokenizer to form a multimodal prompt: $ \hat{p} = [\hat{w}_c] [T_c]$. In doing so, we can leverage the ViL model's text transformer $\mathbb{T}()$, denoted as
\begin{align}
    \hat{z}_{c} = \mathbb{T}(\hat{p}) \in \mathbb{R}^{C \times D},
\end{align}
to obtain the multi-modal representation $\hat{z}_{c}$ with both visual (support videos) and textual (class names) modalities jointly encoded for action class $c$.}
%

\color{black}{For TAD, a background class is needed which however is lacking from
the vocabulary of ViL model.
To solve this, we learn a specific background embedding $\hat{z}_{bg} \in \mathbb{R}^{D}$ with random initialization.
We append this to 
$\hat{z}_{c}$, yielding a complete multimodal representation $E_{mm} \in \mathbb{R}^{(C+1) \times D}$.}

\subsection{Query Feature Regulator}
\label{sec:query_reg}
\begin{figure}[t]
    \centering
    \includegraphics[scale=0.20]{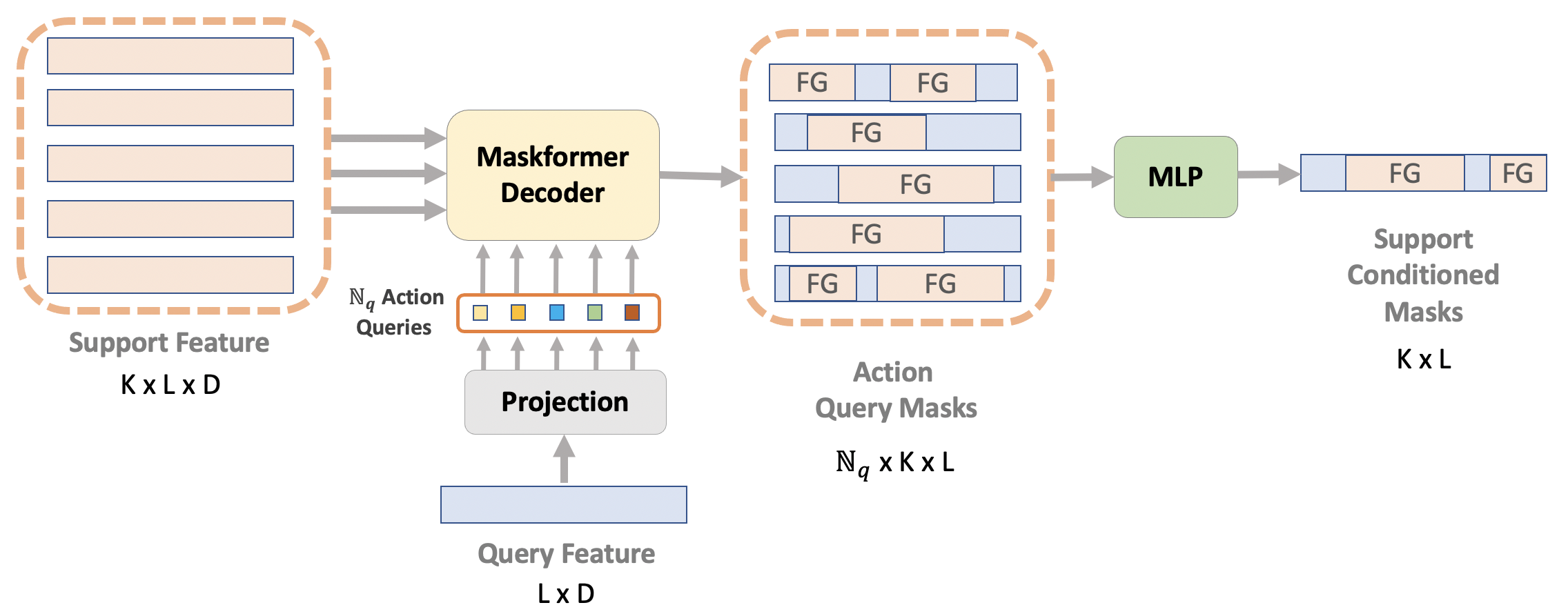}
    \caption{Support conditioned representation masking.}
    \label{fig:rep_mask}
\end{figure}

\textcolor{black}{As shown in \FX{the middle of} Fig.~\ref{fig:overview}, 
\FX{we also design a transformer-based query regulator in parallel with multi-model prompt learning.}
\FX{This regulator takes as input the features from support video and query video, and it }
facilitates the association of action instances across support and query videos in the same action class, with typically large differences (\ie, large intra-class variation).
\FX{Inspired by representation masking \cite{nag2022pclfm} for suppressing background, we also introduce a representation masking mechanism as shown in Fig.~\ref{fig:rep_mask}, which uses the support masked feature for query adaptation.}
}

\noindent \textbf{Representation masking} \textcolor{black}{
We start with this masking using a transformer decoder $\mathcal{S}$. 
Given per-class temporal features of a query video $E_{q} \in \mathbb{R}^{ D \times L}$, we project that to $\mathbb{N}_{q}$ query embeddings $\mathcal{Q}$. 
Cross-attended with the support video features $E_{s} \in \mathbb{R}^{K \times D \times L}$, we generate $\mathbb{N}_{q}$ latent embeddings, followed by a mask projection to obtain $H$ as:
\begin{align}
    H = \mathcal{S}(E_{s},\mathcal{Q}; \theta_{m}) \in \mathbb{R}^{ \mathbb{N}_{q} \times K \times L},
\end{align}
where $\theta_{m}$ are the trainable parameters of $\mathcal{S}$. $H$ represents the most-probable foreground masks in the support videos. To obtain a single mask per location $\hat{H} \in \mathbb{R}^{K \times L}$, 
a tiny MLP for 1D masking is employed.
This support video mask is then binarized yielding the foreground mask $\hat{H}_{bin}$.
The support masked representation $E^{fg}_{s}$ is obtained by applying $\hat{H}_{bin}$ on $E_{s}$.
This masking is trained on the support videos with ground truth and applied to the query videos to obtain the support masked features.
}

\noindent {\bf Query feature regulation }
\textcolor{black}{We use the support masked feature for regularizing
the query feature by cross-attention.
Specifically, with a transformer encoder $\mathcal{C}$, we set the query video feature as its query $\mathbb{Q}$, and the support masked feature as its key $\mathbb{K}$ and value $\mathbb{V}$.
As the number of support videos per class varies,
we aggregate $\mathbb{K}$ and $\mathbb{V}$ by averaging over the number of shots to match a query video.  We then concatenate $\mathbb{K}$/$\mathbb{V}$  with the query feature to form an enhanced version
$\mathbb{K}_{agg}$ and $\mathbb{V}_{agg}$.
The query feature is finally regulated via $\overline{E}_{q} = \mathcal{C}(E_{q},\mathbb{K}_{agg} , \mathbb{V}_{agg}; \theta_{q})$ where $\theta_{q}$ is a learnable parameter.}

\subsection{TAD Classifier and Localizer Heads}
We adopt the TAD head design of \cite{nag2022pclfm,nag2022gsm}
with parallel classification and mask prediction. 

\noindent{\bf Multimodal classifier }
The output of the query feature regulator is fed to the classifier head. We exploit $\hat{E}_{mm} \in \mathbb{R}^{(C+1) \times D}$
as a multimodal classifier to classify the
regulated query features $\overline{E}_{q} \in \mathbb{R}^{L \times D}$ as:
\begin{align}
    \mathcal{P} = \rho((\hat{E}_{mm}*(\overline{E}_{q})^{T})/ \tau) \in \mathbb{R}^{(C+1) \times L},
\end{align}
where each column of $\mathcal{P}$ is the classification result $p_{l} \in \mathbb{R}^{(C+1) \times 1}$  of each snippet $t \in L$,
$\tau=0.7$ is a temperature coefficient and $\rho$ denotes the softmax function.



\noindent\textbf{Action mask localizer} In parallel to classification, this stream predicts 1D binary masks of action instances over the whole video. We use a stack of 1D dynamic-convolution layers to form the mask classifier $\mathbb{H}$. Specifically, given $t$-th snippet $\overline{E}_{q}(t)$, it outputs a 1D mask $m_{t} = [q_{1},...,q_{L}] \in \mathbb{R}^{L \times 1}$ with each $q_{i} \in [0,1] (i \in [1,L])$ meaning the action probability at $i$-th snippet. We define it formally: 
\begin{align}
    \mathcal{M} = \sigma(\mathbb{H}(E_{q}), \theta_{l}),
\end{align}
where $\sigma$ is a sigmoid activation and $t$-th column of $\mathcal{M}$ is the mask prediction by $t$-th snippet and $\theta_{l}$ is the learnable parameter.

\subsection{Meta Training and Inference}

\begin{figure}
    \centering
    \includegraphics[scale=0.18]{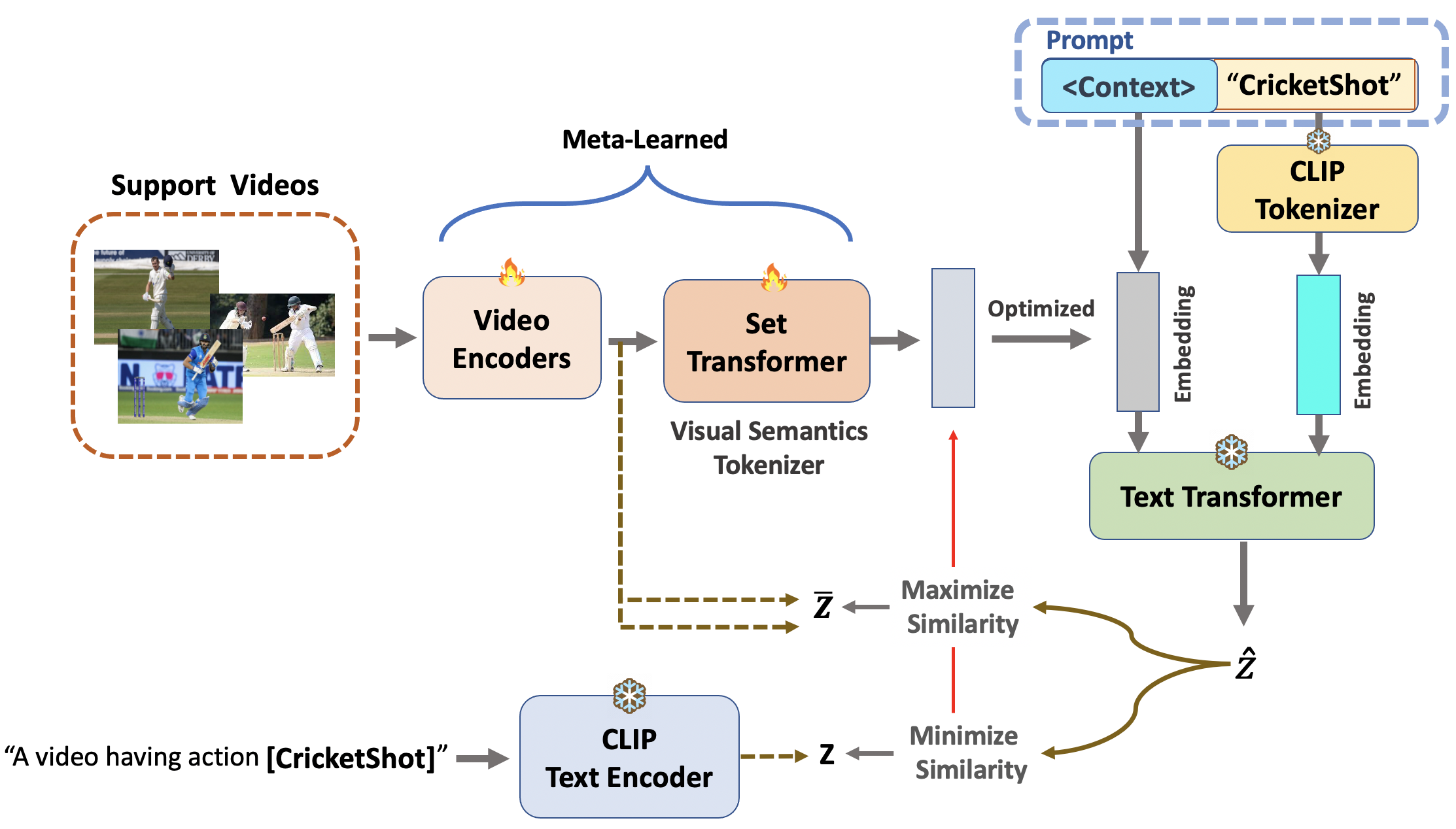}
    \caption{\textbf{Multi-modal prompt optimization}: Given a set of support video examples of a
action class and its
label (``Cricket Shot''), we embed them
with the meta-learned video encoder and predict an embedding using the visual semantics tokenizer $f_{\theta_{s}}$.
We then further tune the
embedding with a contrastive loss $\mathcal{L}_{tok}$}
    \label{fig:optim}
      \vspace{-0.25in}
\end{figure}


\begin{table*}[t]
\centering
\small
\setlength{\tabcolsep}{8pt}

\begin{tabular}{cc|c|cc|cccc|cccc}
\hline
\multicolumn{2}{c|}{}                                        &                                  & \multicolumn{2}{c|}{\textbf{Modality}}                              & \multicolumn{4}{c|}{\textbf{ActivityNetv1.3}}                                                                                                                                         & \multicolumn{4}{c}{\textbf{THUMOS14}}                                                                                                                                                \\ \cline{4-13} 
\multicolumn{2}{c|}{\multirow{-2}{*}{\textbf{Method}}}       & \multirow{-2}{*}{\textbf{N-way}} & \multicolumn{1}{c|}{\textbf{Visual}}       & \textbf{Text}          & \textbf{0.5}                          & \textbf{0.75}                         & \multicolumn{1}{c|}{\textbf{0.95}}                        & \textbf{Avg}                          & \textbf{0.3}                          & \textbf{0.5}                          & \multicolumn{1}{c|}{\textbf{0.7}}                          & \textbf{Avg}                          \\ \hline
\multicolumn{1}{c|}{}                        & FS-Trans      &                                  & \multicolumn{1}{c|}{}                      &                        & 42.2                                  & 24.8                                  & \multicolumn{1}{c|}{5.2}                                 & 25.6                                  & 42.6                                  & 25.7                                  & \multicolumn{1}{c|}{8.2}                                   & 25.5                                  \\
\multicolumn{1}{c|}{\multirow{5}{*}{FS}}    & QAT           & \multirow{-1}{*}{1}              & \multicolumn{1}{c|}{\multirow{5}{*}{\xmark}}  & \multirow{5}{*}{\xmark}   & 44.6                                  & 26.4                                  & \multicolumn{1}{c|}{4.9}                                  & 26.9                                  & 38.7                                  & 24.4                                  & \multicolumn{1}{c|}{7.5}                                   & 24.3                                  \\ \cline{2-2}
\multicolumn{1}{c|}{\multirow{3}{*}{}}    & \textbf{MUPPET}           & \multirow{-2}{*}{}              & \multicolumn{1}{c|}{\multirow{3}{*}{}}  &   & \cellcolor[HTML]{EFEFEF} \textbf{45.4}                                  & \cellcolor[HTML]{EFEFEF} \textbf{28.1}                                  & \multicolumn{1}{c|}{\cellcolor[HTML]{EFEFEF} \textbf{5.6} }                                  & \cellcolor[HTML]{EFEFEF} \textbf{27.8}                                  & \cellcolor[HTML]{EFEFEF} \textbf{44.1}                                  & \cellcolor[HTML]{EFEFEF} \textbf{26.2}                                  & \multicolumn{1}{c|}{\cellcolor[HTML]{EFEFEF} \textbf{8.5}}                                   & \cellcolor[HTML]{EFEFEF} \textbf{26.1}                                  \\\cline{6-13}
\cline{2-3}
\multicolumn{1}{c|}{}                        & Feat-RW       &                                  & \multicolumn{1}{c|}{}                      &                        & 30.7                                  & 16.6                                  & \multicolumn{1}{c|}{2.9}                                  & 17.1                                  & 35.3                                  & 19.6                                  & \multicolumn{1}{c|}{6.8}                                   & 20.1                                  \\
\multicolumn{1}{c|}{}                        & Meta-DETR     &                                  & \multicolumn{1}{c|}{}                      &                        & 32.9                                  & 20.3                                  & \multicolumn{1}{c|}{4.6}                                  & 19.4                                  & 37.5                                  & 20.7                                  & \multicolumn{1}{c|}{7.5}                                   & 21.9                                  \\
\multicolumn{1}{c|}{}                        & FSVOD         & \multirow{-2}{*}{5}              & \multicolumn{1}{c|}{}                      &                        & 34.5                                    & 18.9                                    & \multicolumn{1}{c|}{5.1}                                   & 21.6                                    & 37.9                                  & 23.8                                  & \multicolumn{1}{c|}{7.3}                                   & 22.8                                  \\ \cline{2-2}
\multicolumn{1}{c|}{}                        & \textbf{MUPPET}         &               & \multicolumn{1}{c|}{}                      &                        & \cellcolor[HTML]{E4C9C9} \textbf{36.9}                                    & \cellcolor[HTML]{E4C9C9} \textbf{22.2}                                    & \multicolumn{1}{c|}{\cellcolor[HTML]{E4C9C9} \textbf{5.9}}                                   & \cellcolor[HTML]{E4C9C9} $\textbf{23.0}$                                    & \cellcolor[HTML]{E4C9C9} \textbf{41.2}                                  & \cellcolor[HTML]{E4C9C9} \textbf{25.7}                                  & \multicolumn{1}{c|}{\cellcolor[HTML]{E4C9C9} \textbf{8.5}}                                   & \cellcolor[HTML]{E4C9C9} \textbf{24.9}                                  \\\cline{2-3} \cline{6-13} 
 \hline
\multicolumn{1}{c|}{}                        & OV-DETR     &                                  & \multicolumn{1}{c|}{}                      &                        & 44.2                                  & 27.9                                  & \multicolumn{1}{c|}{6.3}                                  & 28.7                                  & 46.1                                  & 29.7                                  & \multicolumn{1}{c|}{9.0}                                   & 30.4                                  \\
\multicolumn{1}{c|}{}                        & Owl-Vit       &                                  & \multicolumn{1}{c|}{}                      &                        & 43.7                                  & 27.0                                  & \multicolumn{1}{c|}{6.0}                                  & 27.2                                  & 45.2                                  & 29.0                                  & \multicolumn{1}{c|}{9.0}                                   & 30.2                                  \\
\multicolumn{1}{c|}{}                        & EffPrompt        &                                  & \multicolumn{1}{c|}{}                      &                        & 45.9                                  & 27.9                                  & \multicolumn{1}{c|}{5.2}                                  & 29.4                                  & 47.2                                  & 30.4                                  & \multicolumn{1}{c|}{9.8}                                   & 31.1                                  \\
\multicolumn{1}{c|}{}                        & STALE      &                                  & \multicolumn{1}{c|}{\multirow{-4}{*}{\xmark}}  &                        & 47.7                                  & 29.3                                  & \multicolumn{1}{c|}{7.6}                                  & 30.3                                  & 48.9                                  & 32.1                                  & \multicolumn{1}{c|}{10.3}                                  & 32.0                                  \\ \cline{4-4}
\multicolumn{1}{c|}{}                        & Baseline-I        &                                  & \multicolumn{1}{c|}{}                      &                        & 46.9                                   & 28.6                                    & \multicolumn{1}{c|}{6.9}                                   & 29.7                                    & 47.3                                    & 30.5                                    & \multicolumn{1}{c|}{9.2}                                    & 31.8                                    \\ \cline{2-2} \cline{6-13} 
\multicolumn{1}{c|}{}                        & \textbf{\shortmodelname} & \multirow{-6}{*}{1}              & \multicolumn{1}{c|}{\multirow{-2}{*}{\cmark}} &                        & \cellcolor[HTML]{FFE5C6}\textbf{49.7} & \cellcolor[HTML]{FFE5C6}\textbf{32.9} & \multicolumn{1}{c|}{\cellcolor[HTML]{FFE5C6}\textbf{9.2}} & \cellcolor[HTML]{FFE5C6}\textbf{32.7} & \cellcolor[HTML]{FFE5C6}\textbf{50.6} & \cellcolor[HTML]{FFE5C6}\textbf{33.5} & \multicolumn{1}{c|}{\cellcolor[HTML]{FFE5C6}\textbf{11.2}} & \cellcolor[HTML]{FFE5C6}\textbf{33.8} \\ \cline{2-4} \cline{6-13} 
\multicolumn{1}{c|}{}                        & OV-DETR     &                                  & \multicolumn{1}{c|}{}                      &                        & 39.8                                  & 22.3                                  & \multicolumn{1}{c|}{5.4}                                  & 23.1                                  & 40.4                                  & 23.9                                  & \multicolumn{1}{c|}{7.5}                                   & 24.0                                  \\
\multicolumn{1}{c|}{}                        & Owl-Vit       &                                  & \multicolumn{1}{c|}{}                      &                        & 37.9                                  & 20.3                                  & \multicolumn{1}{c|}{5.6}                                  & 21.9                                  & 38.3                                  & 21.9                                  & \multicolumn{1}{c|}{7.7}                                   & 22.6                                  \\
\multicolumn{1}{c|}{}                        & EffPrompt        &                                  & \multicolumn{1}{c|}{}                      &                        & 41.1                                  & 21.6                                  & \multicolumn{1}{c|}{5.4}                                  & 23.8                                  & 39.5                                  & 23.5                                  & \multicolumn{1}{c|}{7.6}                                   & 24.8                                  \\
\multicolumn{1}{c|}{}                        & STALE      &                                  & \multicolumn{1}{c|}{\multirow{-4}{*}{\xmark}}  &                        & 42.3                                  & 22.9                                  & \multicolumn{1}{c|}{6.8}                                  & 24.5                                  & 40.7                                  & 24.9                                  & \multicolumn{1}{c|}{7.1}                                   & 25.4                                  \\ \cline{4-4}
\multicolumn{1}{c|}{}                        & Baseline-I        &                                  & \multicolumn{1}{c|}{}                      &                        & 42.1                                  & 22.7                                  & \multicolumn{1}{c|}{6.0}                                  & 24.0                                  & 40.2                                  & 24.7                                  & \multicolumn{1}{c|}{7.0}                                   & 25.0                                  \\ \cline{2-2} \cline{6-13} 
\multicolumn{1}{c|}{\multirow{-12}{*}{MMFS}} & \textbf{\shortmodelname} & \multirow{-6}{*}{5}              & \multicolumn{1}{c|}{\multirow{-2}{*}{\cmark}} & \multirow{-12}{*}{\cmark} & \cellcolor[HTML]{FEFED8}\textbf{45.3} & \cellcolor[HTML]{FEFED8}\textbf{25.6} & \multicolumn{1}{c|}{\cellcolor[HTML]{FEFED8}\textbf{6.3}} & \cellcolor[HTML]{FEFED8}\textbf{26.2} & \cellcolor[HTML]{FEFED8}\textbf{42.3} & \cellcolor[HTML]{FEFED8}\textbf{27.2} & \multicolumn{1}{c|}{\cellcolor[HTML]{FEFED8}\textbf{7.8}}  & \cellcolor[HTML]{FEFED8}\textbf{27.5} \\ \hline
\multicolumn{1}{c|}{}                        & EffPrompt     &                                  & \multicolumn{1}{c|}{\xmark}                    &                        & 32.0                                   & 19.3                                   & \multicolumn{1}{c|}{2.9}                                  & 19.6                                   & 37.2                                   & 21.6                                   & \multicolumn{1}{c|}{7.2}                                   & 21.9                                   \\ \cline{4-4}
\multicolumn{1}{c|}{}                        & STALE         &                                  & \multicolumn{1}{c|}{\xmark}                    &                        & 32.1                                   & 20.7                                   & \multicolumn{1}{c|}{5.9}                                  & 20.5                                   & 38.3                                   & 21.2                                   & \multicolumn{1}{c|}{7.0}                                   & 22.2                                   \\ \cline{4-4}
\multicolumn{1}{c|}{}                        & Baseline-I        &                                  & \multicolumn{1}{c|}{\cmark}                   &                        & 30.6                                   & 18.0                                   & \multicolumn{1}{c|}{4.1}                                  & 18.7                                   & 35.8                                   & 20.5                                   & \multicolumn{1}{c|}{7.1}                                   & 20.8                                   \\ \cline{2-2} \cline{4-4} \cline{6-13} 
\multicolumn{1}{c|}{\multirow{-4}{*}{ZS}}    & \textbf{\shortmodelname} & \multirow{-4}{*}{All}            & \multicolumn{1}{c|}{\xmark}                    & \multirow{-4}{*}{\cmark}  & \cellcolor[HTML]{DAE8FC}\textbf{33.5}  & \cellcolor[HTML]{DAE8FC}\textbf{21.9}  & \multicolumn{1}{c|}{\cellcolor[HTML]{DAE8FC}\textbf{6.7}} & \cellcolor[HTML]{DAE8FC}\textbf{22.0}  & \cellcolor[HTML]{DAE8FC}\textbf{40.1}  & \cellcolor[HTML]{DAE8FC}\textbf{22.8}  & \multicolumn{1}{c|}{\cellcolor[HTML]{DAE8FC}\textbf{8.1}}  & \cellcolor[HTML]{DAE8FC}\textbf{24.8}  \\ \hline
\end{tabular}
\caption{Comparing our \shortmodelname{} with prior art few-shot (FS), zero-shot (ZS) and alternative methods.
    {\em Setting}: 5-shot; the CLIP model for multimodal few-shot (MMFS) methods; 50\%/50\% train/test class split for all ZS methods.}
    \label{tab:main}
    \vspace{-0.1in}
\end{table*}

\noindent{\bf Learning objective} 
Following \cite{nag2022pclfm},
we adopt the cross-entropy loss $\mathcal{L}_{c}$ for classification,  binary dice loss $\mathcal{L}_{m}$ and binary mask loss $\mathcal{L}_{comp}$ for masking. 
For more effective training, we further impose a contrastive criterion \cite{chen2020simple} to optimize the visual semantics tokenizer ({Fig. \ref{fig:optim}}). 
Given the multimodal representation $\hat{z}_{c}$, original prompt embedding $z_{c}$, and video embedding $\overline{z}_{c}$ for each class $c$, the contrastive loss is defined as: 
\begin{align}
    \mathcal{L}_{tok} = - \log\frac{\exp(cos(\overline{z}_{c},\hat{z}_{c}))}{\exp(cos(\overline{z}_{c},\hat{z}_{c})) + 2\exp(cos(z_{c},\hat{z}_{c}))},
\end{align}
where $cos(.)$ is cosine similarity and the factor $2$ is for contrasting $z_{c}$ with both visual and textual embeddings. To contrast background ($z_{bg}$) from foreground ($\hat{z}_c$), we minimize:
\begin{align}
    \mathcal{L}_{bg} = argmin \sum^{C}_{j=1}(cos(z_{bg}, z^{j}_{c}) - \delta_{bg})^{2},
\end{align}
where 
$\delta_{bg}$ is the margin hyper-parameter. 

\noindent{\bf Training }
\textcolor{black}{
The training procedure
has two stages. 
In the {\bf\em first} stage, the model is trained on the base dataset
$D_{base}$ in a standard supervised learning manner.
More specifically,  we deploy the objective
$\mathcal{L}_{base} = \mathcal{L}_{c} + \mathcal{L}_{m} + \mathcal{L}_{comp} + \mathcal{L}_{tok} + \mathcal{L}_{bg} + \mathcal{L}_{const}$ and optimize the parameter set $\phi_{base} = \{ \theta,\phi,\theta_{s}, \theta_{m},\theta_{q}, \theta_{l}\}$. This trains all modules of \shortmodelname{} except the language model end-to-end. 
The {\bf\em second} stage is for few-shot fine-tuning,
following a typical meta-learning paradigm.
Due to no ground-truth for query videos under this setting,
we freeze the query regularizer ($\theta^{base}_{q}$) and action localizer head ($\theta^{base}_{l}$).
We thus minimize $\phi_{meta} = \{ \theta,\phi,\theta_{s}, \theta_{m}\}$, with 
$\mathcal{L}_{m}$ and $\mathcal{L}_{c}$ removed.
More details on meta-learning is given in \texttt{Supplementary}.
}

\noindent{\bf Inference } At test time, we generate action instance predictions for each query video by the classification $\bm{P}$ and mask $\bm{M}$ predictions following \cite{nag2022pclfm}.
For each top scoring action snippet in $\bm{P}$,
we then obtain the
temporal masks by thresholding the corresponding column of $\bm{M}$ using a set of thresholds $\Theta=\{\theta_i\}$.
We apply SoftNMS \cite{bodla2017soft} to obtain the final top-scoring outputs.

\section{Experiments}


\noindent \textbf{Datasets } We evaluate two popular TAD benchmarks.
(1) ActivityNet-v1.3 \cite{caba2015activitynet} has 19,994 videos from 200 action classes. We follow the
standard split setting of 2:1:1 for train/val/test.
(2) THUMOS14 \cite{idrees2017thumos} has 200 validation videos and 213 testing
videos from 20 categories with labeled temporal boundary and action class.

\noindent\textbf{Settings}
We consider two major settings. 
\texttt{Few-shot setting}: To facilitate fair comparison, we adopt the same dataset and class split as \cite{nag2021few}. For both datasets,
we divide all the classes into three non-overlapping subsets for training (80\%), validation (10\%), and testing (10\%), respectively. 
The validation set is used for model parameter tuning and 
best model selection. We consider 1-way/class and 5-way settings.
We consider naturally untrimmed support videos. For each $N$-way $K$-shot experiment, we divide the base and novel class videos into few-shot episodes where each episode consists of $N \times (K+1)$ tasks. We train with 1000 episodes and test with 250 episodes with random tasks and report the average result.
\texttt{Zero-shot setting}: In this setting, similar to few-shot, we ensure that $D_{val} \bigcap D_{test} = \phi$. We follow the setting and dataset splits used by \cite{nag2022pclfm} for a fair comparison. For both datasets, we train with $50\%$ classes and test on $50\%$ classes. To ensure statistical significance, we conduct 10 random samplings to split classes for each setting, following \cite{ju2021prompting}. More details on splits are provided in \texttt{Supplementary}.

\noindent\textbf{Implementation details }
For a fair comparison, we use CLIP \cite{radford2021learning} initialized weights for both datasets. 
For comparing with CLIP-based TAD baselines, we use the image and text encoders from pretrained CLIP (ViT-B/16+Transformer) \cite{radford2021learning}. We also used Kinetics \cite{kay2017kinetics} pretrained initialization for showing the robustness of our approach.
Video frames are pre-processed to $112 \times 112$ in spatial resolution, and the maximum number of textual tokens is $77$, following CLIP. Given a variable-length video, we first sample every 6 consecutive frames as a snippet. Then we feed the snippet
into our vision encoder and extract the features before the
fully connected layer. Thus, we obtain a sequence of snippet-level
feature for the untrimmed video. After this, each video’s feature sequence $F$ is rescaled to $T = 100/256$ snippets for ActivityNet/THUMOS using linear interpolation. Our model is trained on 6 NVIDIA 3090RTX GPUs with 1000/250 episodes using Adam optimizer with a learning rate of {$10^{-4}/10^{-5}$ for ActivityNet/THUMOS respectively during base and meta-training. More implementation details are provided in \texttt{Supplementary}}.

\vspace{-0.05in}
\subsection{Comparison with state-of-the-art}

\begin{figure*}[t]
    \centering
    \includegraphics[scale=0.40]{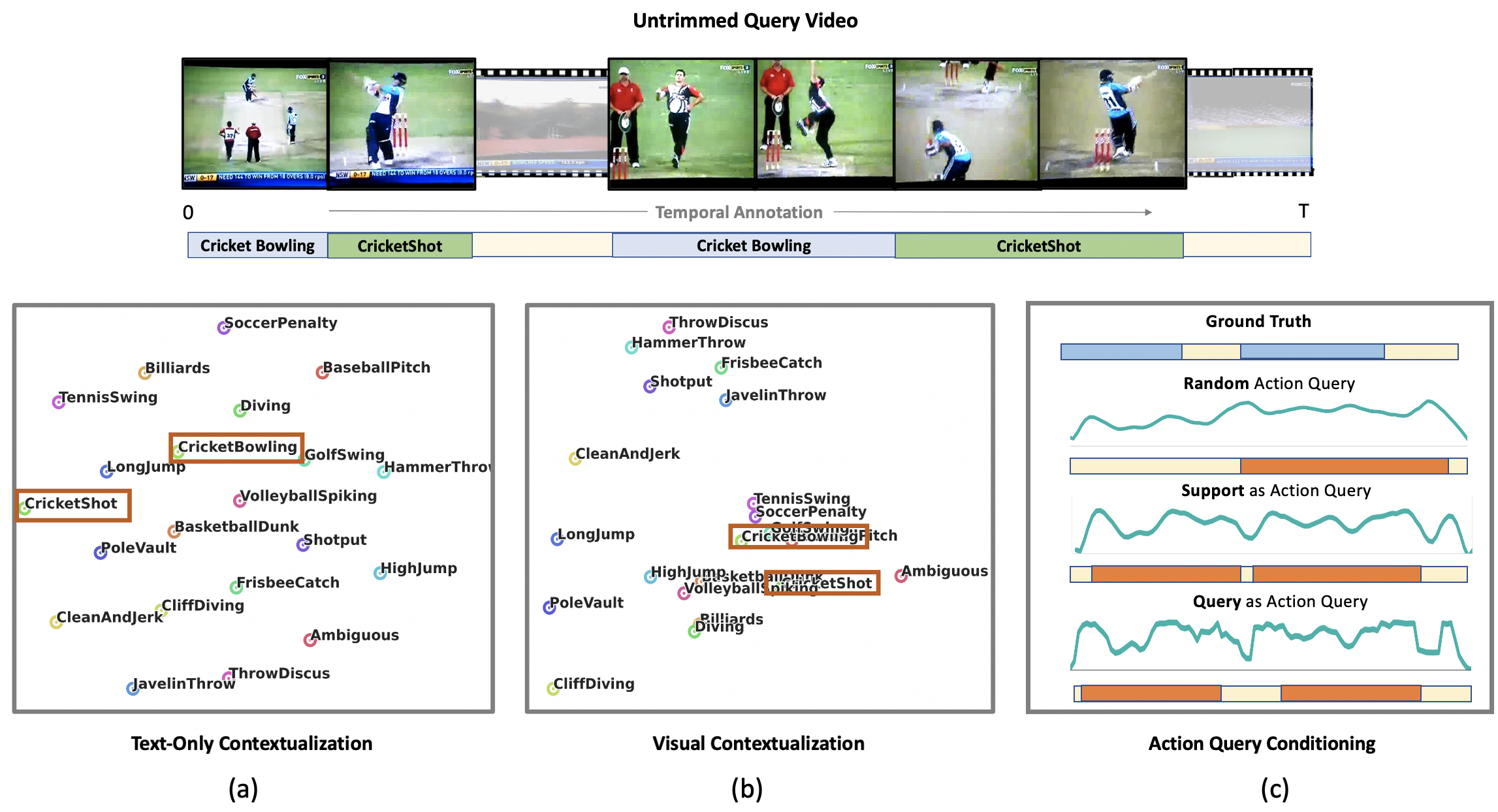}
    \caption{\textbf{Illustration of the impact of \shortmodelname{} on a random video} (a) PCA plot of our model with textual prompts (b) PCA plot of our model after incorporating visual semantics (c) Impact of various action query initialization method on actionness of representation mask.}
    \label{fig:exp_vis}
\end{figure*}

\noindent \textbf{Competitors }
We consider extensively three sets of previous possible methods:
{\bf(1)} {\bf\em Few-shot learning based} methods: 
Two action detection methods (FS-Trans \cite{yang2021few} and QAT \cite{nag2021few}). 
AS FS-Trans was originally designed for spatiotemporal action detection, we discarded the spatial detection part here. 
Due to limited FS-TAD models, we adapt 2 object detection baselines (Feat-RW \cite{kang2019few}, Meta-DETR \cite{zhang2021meta}). We replaced their backbones with CLIP ViT encoders and the object decoders with TAD decoders. 
We similarly adapted a video-based object detection method (FSVOD \cite{fan2021few}) where temporal action proposals and temporal matching network are applied with TAD decoder. For a fair comparison, we deploy \shortmodelname{} in the FS setting by discarding the textual input.  
%
{\bf(2)} {\bf\em Multi-modal Few-shot learning based} methods:
As this is a new problem, we need to adapt existing methods for baselines.
We adapted \textcolor{black}{zero-shot} object detection methods (OV-DETR \cite{zang2022open}, OWL-ViT \cite{minderer2022simple}) \textcolor{black}{as they can facilitate the multi-modal setting due to their CLIP based design}.
\textcolor{black}{For \cite{zang2022open}, we kept the encoder unchanged for frame-level extraction and replaced the decoder with a start/end regressor as RTD-Net \cite{Tan_2021_RTD}. For \cite{minderer2022simple}, we used the encoder backbone unchanged and replaced the bounding box detectors with a start/end regressor as BMN \cite{lin2019bmn}.}
%
We also considered two TAD methods (EffPrompt \cite{ju2021prompting} and STALE \cite{nag2022pclfm}) by finetuning all modules with support set during inference. 
%
We further adapted \textcolor{black}{zero-shot classification method} CoCOOP \cite{zhou2022conditional} (denoted as \texttt{Baseline-I}) to zero-shot TAD model STALE \cite{nag2022pclfm} by adding the meta-network from visual branch to learn the textual tokens.
%
This is the closest competitor of our proposed \shortmodelname.
{\bf(3)} {\bf\em Zero-shot learning based} methods: EffPrompt \cite{ju2021prompting} and STALE \cite{nag2022pclfm}
and \texttt{Baseline-I}.
We deploy \shortmodelname{} in ZS setting by discarding the FS components (\eg, visual-semantics tokenizer and query regularizer).
%
All the methods use the same CLIP \textcolor{black}{ViT \cite{radford2021learning} vision encoders} for a fair comparison.

\noindent \textbf{Results }
We make several observations from the results in Table~\ref{tab:main}. 
{\bf (1)} \textbf{\em FS setting}: Even with \textcolor{black}{1-way} 
support sets, FS-TAD methods (FS-Trans \cite{yang2021few}, QAT \cite{nag2021few}) still outperform
5-way 
object detection based counterparts (Feat-RW \cite{kang2019few}, Meta-DETR \cite{zhang2021meta}, FSVOD \cite{fan2021few}). 
This indicates the importance of modeling temporal dynamics and task-specific design. \textcolor{black}{Our \shortmodelname{} outperforms the 1/5-way alternatives by 0.9/1.4\% margin verifying the superiority of our model design.}
{\bf (2)} \textbf{\em MMFS setting}: Interestingly, object detection methods (OV-DETR \cite{zang2022open}, OWL-ViT \cite{minderer2022simple}) can perform similarly as FS-TAD (EffPrompt \cite{ju2021prompting}, STALE \cite{nag2022pclfm}) ones when using text modality. Our Baseline-I yields competitive performance. Notably, \shortmodelname{} surpasses the best FS-TAD model (QAT) by a margin of $5.8 \%$, validating the superiority of our model and our motivation for MMFS-TAD.
\textcolor{black}{In particular, QAT tackles 1-way FS-TAD (\ie, foreground class vs. background)
similar to action proposal generation. 
The result suggests that 
our multimodal classifier is better than 
the popular UntrimmedNet.}
A similar observation can be drawn in the 5-way case. 
{\bf (3)} \textbf{\em ZS setting}: 
Our \shortmodelname{} is superior to recent art models (EffPrompt \cite{ju2021prompting}, STALE \cite{nag2022pclfm}) and \texttt{Baseline-I} (an integrated model even using training videos).
This verifies the flexibility of our method in deployment,
in addition to promising performance.

\subsection{Ablation Studies}

\paragraph{MMFS {\em vs.} FS setting} \textcolor{black}{MUPPET by the design choice can work in both few-shot setting (by removing the language encoder) and multi-modal few-shot setting. 
To examine the usefulness of textual semantic information,
we compare the result under the 5-way 5-shot setting on ActivityNet.
As shown in Table \ref{tab:mmfs_vs_fs}, 
with \shortmodelname{} a gain of 3.2\% can be benefited 
from the class semantic description.
%
This validates our motivation that text modality can compensate for the limited few-shot examples.}

\begin{table}[h]
\centering
\small
\caption{MUPPET in 5-way 5-shot setting on ActivityNet.}
\begin{tabular}{c|c|ccc|c}
\hline
\bf{Setting} & \bf{Text} & \bf{0.5}  & \bf{0.75} & \bf{0.95} & \bf{Avg}  \\ \hline
FS      & \xmark   & 36.9 & 22.2 & 5.9  & 23.0 \\ \hline
MMFS    & \cmark  & \textbf{45.3} & \textbf{25.6} & \textbf{6.3}  & \textbf{26.2} \\ \hline
\end{tabular}
\label{tab:mmfs_vs_fs}
\end{table}

\begin{table}[h]
\footnotesize
    \setlength{\tabcolsep}{6pt}
 \centering
 \caption{Prompt learning design on ActivityNet.
 Setting: 5-way.}
\label{tab:prompt}
\begin{tabular}{@{}c|c|cc|cc@{}}
\toprule
\multirow{2}{*}{Design} & \multirow{2}{*}{ Shots} & \multicolumn{2}{c|}{Prompt style} & \multicolumn{2}{c}{mAP}   \\ \cmidrule(l){3-6} 
                              &                               & Learnable      & Context          & 0.5         & Avg         \\ \midrule
LPS                    & -                             & \xmark             & -                & 18.4          & 13.6          \\ \midrule
LVP                    & 5                             & \cmark             & Visual                & 43.2          & 25.0          \\ \midrule
LTP                    & 5                          & \cmark            & Text         & 42.7          & 24.7          \\
\textbf{Ours}            & \textbf{1}                 & \textbf{\cmark}   & \textbf{Visual}  & \textbf{43.7} & \textbf{25.1} \\
\textbf{Ours}            & \textbf{5}                 & \textbf{\cmark}   & \textbf{Visual}  & \textbf{45.3} & \textbf{26.2} \\
\bottomrule
\end{tabular}
\end{table}

\begin{figure}[t]
    \centering
    \includegraphics[width=3.3in]{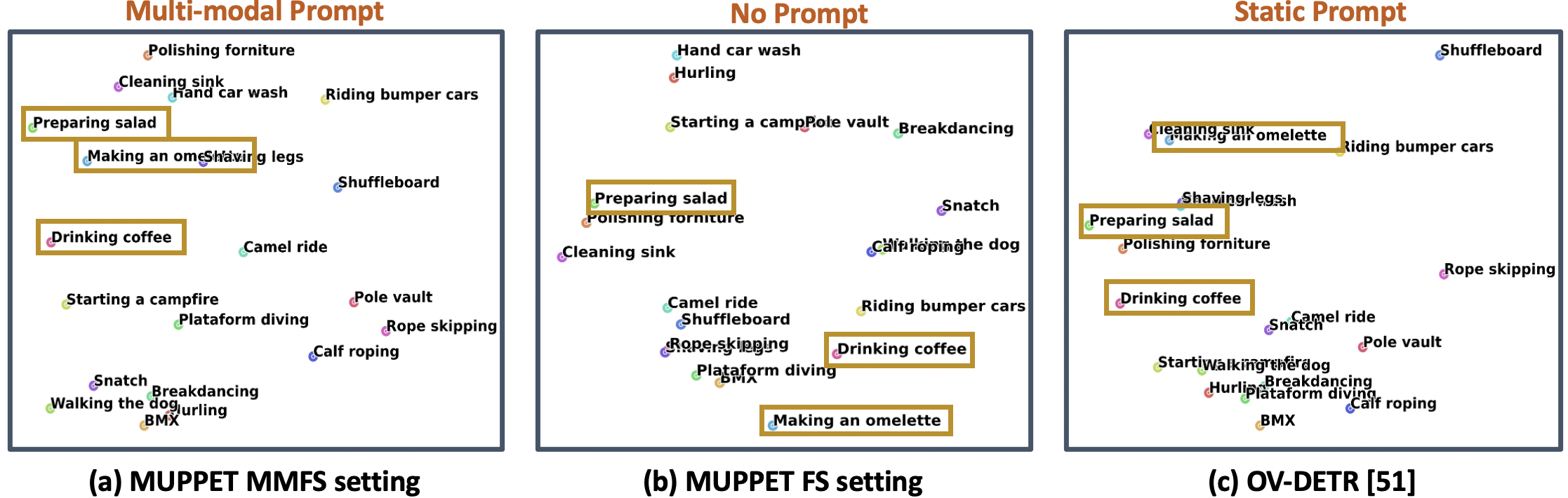}
    \vspace{-0.15in}
    \caption{PCA of classifier weights on ActivityNet.
    As highlighted in the boxes, it is evident that visual information is useful in grouping related actions such as ``preparing salad'' and ``making an omelet'' so that the embedding space is made more meaningful.
    Best viewed when zoom-in.}
    \label{fig:pca}
    \vspace{-0.15in}
\end{figure}

\noindent\textbf{Prompt learning design } 
We evaluate our multimodal prompt meta-learning that meta-learns the semantic information from few-shot support videos.
We compare it against three alternatives:
(i) \textit{Learnable Prompt from Scratch} ({\bf LPS}): Learning the prompt from random vectors without the text encoder of ViL model (CLIP \cite{radford2021learning} in this case).
(ii) \textit{Learnable Textual Prompt} ({\bf LTP}): Learning the prompt from randomly initialized vectors with the text encoder of ViL model.
(iii) \textit{Learnable Visual Prompt} ({\bf LVP}): Learning the prompt from vectors initialized by visual features from the visual encoder of ViL model,
as \texttt{Baseline-I}.
We observe from Table \ref{tab:prompt} that:
{\bf (1)} Leveraging the pretrained text encoder is critical due to \textcolor{black}{pretraining on vast training data},
otherwise, a big drop would occur as demonstrated by LPS.
{\bf (2)} \textcolor{black}{Learning from only few visual samples via a meta-network, 
LVP is inferior to our MUPPET, 
verifying the complementary effect of 
our semantics tokenzier.}
%
{\bf (3)} However, using only text modality for prompt learning (\ie, LTP)
is even inferior to visual modality only (\ie, LVP). This is not surprising 
as videos provide more comprehensive and finer information about new classes. This effect is indicated in Fig.~\ref{fig:pca} that visual information helps in grouping similar actions such as ``making an omelet'' and ``preparing salad''.
%
{\bf (4)} Also, the inferiority of LTP and LVP to ours
suggests that learning class-specific tokens as we design is more suitable 
than learning a set of global prompts shared for all classes in MMFS-TAD.

We further examine the network choice (1D CNN {\em vs.} set Transformer \cite{lee2019set}) for visual semantics tokenizer,
and the necessity of class-specific prompts.
As shown in Table \ref{tab:context}, we see that:
{\bf (1)} A permutation invariant set Transformer is better than 1D CNN.
{\bf (2)} Using a single token per class is enough by our prompting method.
This is different from previous prompting methods \cite{zhou2021learning} 
that instead learn multiple (\eg, 20) global tokens shared by all classes. 

\noindent\textbf{Episodic adapters in video encoder }
We exploit episodic adapters for the video encoder of ViL model.
Alternative methods include (i) {\em Freezing video encoder} without any task adaptation as STALE \cite{nag2022pclfm}, (ii) {\em Fine-tuning} the video encoder.
We also compare with the adapted STALE for MMFS-TAD.
We observe from Table \ref{tab:enc} that:
{\bf(1)} Fine-tuning is indeed useful as expected, as compared to the frozen encoder. However, it is less effective due to limited training data.
{\bf(2)} Using our adapters is the best which alleviates overfitting with higher efficiency by only learning a fraction of the parameters.
\begin{table}[h]
\vspace{-0.1in}
\footnotesize
    \centering
    \setlength{\tabcolsep}{3pt}
 \caption{Design of visual semantics tokenizer on ActivityNet.
 Setting: 5-way 5-shot. 
\#T/C: Tokens per Class.}

\begin{tabular}{c|c|c|cc}
\hline
\multirow{2}{*}{\textbf{Network}} & \multirow{2}{*}{\textbf{Meta-Learn}} & \multirow{2}{*}{\textbf{ \#T/C}} & \multicolumn{2}{c}{\textbf{mAP}}                 \\ \cline{4-5} 
                                &                                      &                                         & \multicolumn{1}{c|}{\textbf{0.5}} & \textbf{Avg} \\ \hline
\multirow{3}{*}{1D CNN}         & \xmark                                   & 20                        & \multicolumn{1}{c|}{37.4}           & 21.3           \\
                                & \cmark                                  & 20                        & \multicolumn{1}{c|}{40.8}           & 23.0           \\
                                & \cmark                                  & 1                                       & \multicolumn{1}{c|}{39.7}           & 22.5           \\ \hline
\multirow{3}{*}{Set Transformer \cite{lee2019set}}        & \xmark                                   & 1                                       & \multicolumn{1}{c|}{43.8}           & 24.7           \\
                                & \cmark                                  & 1                                       & \multicolumn{1}{c|}{\bf{45.3}}           & \bf{26.2}           \\
                                & \cmark                                  & 20                       & \multicolumn{1}{c|}{44.7}           & 25.6          \\ \hline
\end{tabular}
\label{tab:context}
  \vspace{-0.1in}
\end{table}

\noindent \textbf{Query feature regulation } 
MMFS-TAD often presents a large intra-class variation
due to limited training video data.
Our query feature regulation is designed for overcoming this challenge.
As shown in Table \ref{tab:query}, this scheme is effective
with the gain increasing along with the shots of the training set.
This validates the usefulness of our design.
%
In Table \ref{tab:featmask} we observe a gain of 1.3\% in mAP@0.5. 
We also show that
randomly initialization leads to inferior foreground prediction (see Fig~\ref{fig:exp_vis}(c)).
Support video features based initialization 
can improve but still not as strong as query video features as in our design.




\begin{table}[]
\footnotesize
\centering
\setlength{\tabcolsep}{3pt}
         \caption{Video encoder on ActivityNet. Setting: 5-way 5-shot.}
      \centering
      \setlength{\tabcolsep}{8pt}
\begin{tabular}{@{}c|c|cc@{}}
\toprule
\multirow{2}{*}{Method} & {Video} & \multicolumn{2}{c}{mAP}        \\ \cmidrule(l){3-4} 
                        &  encoder                         & \multicolumn{1}{c|}{0.5} & Avg \\ \midrule
\multirow{3}{*}{\shortmodelname}   & Freeze                    & \multicolumn{1}{c|}{41.1}  & 25.3  \\
                        & Full-tuning               & \multicolumn{1}{c|}{45.0}  & 26.1  \\
                        & \textbf{Adapters}                   & \multicolumn{1}{c|}{\textbf{45.3}}  & \textbf{26.2}  \\ \bottomrule
\end{tabular}
\label{tab:enc}
\vspace{-0.05in}
\end{table}

\begin{table}[]
\footnotesize
          \centering
        \caption{Query feature regulation on ActivityNet.
        Setting: 5-way. 
        }
\label{tab:query}
\setlength{\tabcolsep}{3pt}
\begin{tabular}{c|c|cc}
\hline
\multirow{2}{*}{\textbf{$K$-shot}} & \multirow{2}{*}{\textbf{Query Masking}} & \multicolumn{2}{c}{\textbf{mAP}} \\ \cline{3-4} 
                                         &                                             & \textbf{0.5}    & \textbf{Avg}   \\ \hline
-                                       & \xmark                                          & 41.1              & 24.8             \\ \hline
1                                        & \cmark                                         & 43.7              & 25.1             \\
5                                        & \cmark                                         & \bf{45.3}              & \bf{26.2}             \\ \hline
\end{tabular}
\vspace{-0.1in}
\end{table}

\begin{table}[]
\footnotesize
\caption{Representation masking on support video features on ActivityNet.
Setting: 5-way 5-shot.}\label{tab:featmask}
\centering
\begin{tabular}{@{}c|c|cc@{}}
\toprule
\multirow{2}{*}{\textbf{Masking decoder}}     & \multirow{2}{*}{\textbf{Initialization}} & \multicolumn{2}{c}{\textbf{mAP}}                            \\ \cmidrule(l){3-4} 
&                          & \multicolumn{1}{c|}{\textbf{0.5}}           & \textbf{Avg}           \\ \midrule
1D CNN                             & -                        & \multicolumn{1}{c|}{31.7}          & 21.3          \\ \midrule
\multirow{3}{*}{MaskFormer \cite{cheng2021maskformer}} & Random                        & \multicolumn{1}{c|}{38.2}          & 24.9          \\

& Support                      & \multicolumn{1}{c|}{43.7}          & 25.8          \\ 
& \textbf{Query}               & \multicolumn{1}{c|}{\textbf{45.3}} & \textbf{26.2} \\\bottomrule
\end{tabular}
\vspace{-0.15in}
\end{table} 

\paragraph{Ablation of visual adapter:}
We have ablated the role of adapters in feature backbone in Tab~\ref{tab:adapt}. For this experiment, we compared two previous ViT based baselines \textit{Baseline-I} and STALE \cite{nag2022pclfm}. The residual adapters are plugged intot the ViT backbone maintaining the same configuration with MUPPET. From the results in Tab~\ref{tab:adapt}, it is evident that the variant with adapter improves the performance, however, MUPPET w/o adapter variant is still stronger than the adapter infused baselines. This suggests the superiority of our model design.

\begin{table}[h]

\centering
\caption{MUPPET in 5-way 5-shot setting on ActivityNet}
\label{tab:adapt}
\begin{tabular}{c|c|ccc|c}
\hline
\textbf{Method}         & \textbf{Backbone} & \textbf{0.5} & \textbf{0.75} & \textbf{0.95} & \textbf{Avg} \\ \hline
Baseline-1              & w/Adapter         & 42.5               & 23.0               & 6.0               & 24.2              \\
STALE                   & w/ Adapter        & 42.7             & 23.2             & \textbf{6.9}              & 24.7             \\ \hline
\multirow{2}{*}{MUPPET} & w/o Adapter       & 45.0                & 25.2               & 6.0              & 26.1               \\ \cline{2-6} 
                        & w/ Adapter        & \textbf{45.3}              & \textbf{25.6}              & 6.3               & \textbf{26.2}             \\ \hline
\end{tabular}
\end{table}

\noindent \textbf{Ablation with different pretraining}
\textcolor{black}{We experiment our \shortmodelname{} with Kinetics-400 pretraining. Concretely, we use Kinetics-400 pretrained weights for both visual and textual branch provided by ActionCLIP \cite{wang2021actionclip} in this experiment. 
From Table \ref{tab:pretrain} we observe similar findings as that of CLIP \cite{radford2021learning} pretrained features in Table 1 (Main paper). Our {\shortmodelname} outperforms the competitors by similar margin and better than CLIP pretraining by $4\%$ in avg mAP, 
This confirms that the superiority of our method is feature agnostic, with the desired capability of 
reducing the domain gap between pretraining and downstream tasks.
}

\begin{table}[h]
    \centering
     \caption{Analysis of \shortmodelname with different pre-training feature on ActivityNet in 5-way 5-shot setting.}
      \centering
      \setlength{\tabcolsep}{3pt}
\begin{tabular}{c|c|cccc}
\hline
\multirow{2}{*}{\textbf{Method}} & \multirow{2}{*}{\textbf{Feature}} & \multicolumn{4}{c}{\textbf{mAP}}                                                 \\ \cline{3-6} 
                                 &                                          & \textbf{0.5} & \textbf{0.75} & \multicolumn{1}{c|}{\textbf{0.95}} & \textbf{Avg} \\ \hline
EffPrompt                            & CLIP                          & 41.1            & 21.6               & \multicolumn{1}{c|}{5.4  }            & 23.8           \\
STALE                        & CLIP                             & 42.3            & 22.9              & \multicolumn{1}{c|}{6.8}            & 24.5           \\\hline
\textbf{\shortmodelname}                    & \textbf{CLIP}               & \textbf{45.3 }  & \textbf{25.6 }   & \multicolumn{1}{c|}{\textbf{6.3 }}   & \textbf{26.2}  \\
\textbf{\shortmodelname}                    & \textbf{K-400}               & \textbf{48.1}  & \textbf{29.4}   & \multicolumn{1}{c|}{\textbf{10.0}}   & \textbf{30.2}  \\ \hline
\end{tabular}
\label{tab:pretrain}
\end{table}

\subsection{Multi-modal Few-shot Object Detection}
\textcolor{black}{For generality evaluation,
we further adapt \shortmodelname{} to MMFS object detection.
We replace the video-encoder with ViT+adapter \cite{radford2021learning} image encoder and TAD decoder with object classifier and bounding-box regressor. More details can be found in \texttt{Supplementary}.
Experiments are conducted on COCO \cite{lin2014microsoft}, following the same MMFS setup as in TAD.
From Table \ref{tab:obj} it is observed that the multi-modal information is indeed beneficial in few-shot object detection surpassing the nearest competitor 
META-DETR \cite{zhang2021meta} by 0.5\%/1.1\% in 5/10-shot setting.
This suggests that our method can favorably serve as a unified framework for both object and action detection under the multimodal few-shot setting.
}

\begin{table}[h]
\vspace{-0.1in}
\footnotesize
\centering
\setlength{\tabcolsep}{2pt}
\caption{Comparing our adapted \shortmodelname{} with existing few-shot object detection methods on COCO dataset. 
}

\begin{tabular}{c|ccc|ccc}
\hline
                                  & \multicolumn{3}{c|}{\textbf{5-Shot}}                                & \multicolumn{3}{c}{\textbf{10-Shot}}                                \\ \cline{2-7} 
\multirow{-2}{*}{\textbf{Method}} & \textbf{AP}                         & \textbf{$AP_{50}$} & \textbf{$AP_{75}$} & \textbf{AP}                         & \textbf{$AP_{50}$} & \textbf{$AP_{75}$} \\ \hline
FRCN \cite{ren2015faster}                             & \cellcolor[HTML]{38FFF8}4.6         & 8.7           & 4.4           & \cellcolor[HTML]{38FFF8}5.5         & 10.0          & 5.5           \\
TFA w/ cos  \cite{wang2020frustratingly}                      & \cellcolor[HTML]{38FFF8}7.0         & 13.3          & 6.5           & \cellcolor[HTML]{38FFF8}9.1         & 17.1          & 8.8           \\
Deform-DETR  \cite{zhu2020deformable}                     & \cellcolor[HTML]{38FFF8}7.4         & 12.3          & 7.7           & \cellcolor[HTML]{38FFF8}11.7        & 19.6          & 12.1          \\
FSOD  \cite{fan2020few}                            & \cellcolor[HTML]{38FFF8}-           & -             & -             & \cellcolor[HTML]{38FFF8}12.0        & 22.4             & 11.8          \\
QA-FewDet  \cite{han2021query}                      & \cellcolor[HTML]{38FFF8}9.7         & 20.3          & 8.6           & \cellcolor[HTML]{38FFF8}11.6        & 23.9          & 9.8           \\
META-DETR \cite{zhang2021meta}                          & \cellcolor[HTML]{38FFF8}15.4        & 25.0          & \textbf{15.8}          & \cellcolor[HTML]{38FFF8}19.0        & 30.5          & 19.7          \\ \hline
\textbf{MUPPET}                   & \cellcolor[HTML]{38FFF8}\textbf{15.9} & \textbf{26.4}   & 14.8   & \cellcolor[HTML]{38FFF8}\textbf{20.1} & \textbf{32.3}   & \textbf{19.9}   \\ \hline
\end{tabular}
\label{tab:obj}
\end{table}
\vspace{-0.15in}
\section{Conclusions}

In this work, we have presented a
{\em multi-modality few-shot temporal action detection} (MMFS-TAD) problem setting that tackles the conventional FS-TAD and ZS-TAD jointly.
We further proposed a novel {\bf\em MUlti-modality PromPt mETa-learning} (\shortmodelname) method, characterized by prompt meta-learning from multimodal inputs,
adapters-based ViL model adaptation, and query feature regulation.
%
Extensive experiments on two benchmarks
show that our \shortmodelname{} surpasses both strong baselines and state-of-the-art methods under a variety of settings.
\textcolor{black}{We also show the generic superiority of our method in tackling multi-modal few-shot object detection.}


{\small
\bibliographystyle{ieee_fullname}
\bibliography{main_draft}
}

\end{document}